# Mitigating covariate shift in non-colocated data with learned parameter priors

Behraj Khan, *Institute of Business Administration Karachi*
Behroz Mirza, *Habib University Karachi*
Nouman M. Durrani, *National University of Computer and Emergings Sciences, Karachi*
Tahir Syed, *Institute of Business Administration Karachi*

*Abstract*—When training data are distributed across time or space, covariate shift across fragments of training data biases cross-validation, compromising model selection and assessment. We present *Fragmentation-Induced covariate-shift Remediation* (*FIcsR*), which minimizes an $f$-divergence between a fragment's covariate distribution and that of the standard cross-validation baseline. We show an equivalence with popular importance-weighting methods. The method's numerical solution poses a computational challenge owing to the overparametrized nature of a neural network, and we derive a Fisher Information approximation. When accumulated over fragments, this provides a global estimate of the amount of shift remediation thus far needed, and we incorporate that as a prior via the minimization objective. In the paper, we run extensive classification experiments on multiple data classes, over 40 datasets, and with data batched over multiple sequence lengths. We extend the study to the *k*-fold cross-validation setting through a similar set of experiments. An ablation study exposes the method to varying amounts of shift and demonstrates slower degradation with *FIcsR* in place. The results are promising under all these conditions; with improved accuracy against batch and fold state-of-the-art by more than 5% and 10%, respectively.

## I. POSITIONING

**Learning systems and feature evolution** Big data often demand large learning systems that do not presuppose the presence of the complete dataset in one place at the same time which we will refer as non-colocation of data. This implies that a contemporary real-world setting undermines the classical machine learning assumption of data being independently and identically distributed (*iid*), be it among fragments of training data or between training and test or validation sets. Such problems are characterized as distribution shift [30]. The foremost among them is *covarite shift* [7], where training data may have a different feature distribution to test data. In *standard* covariate shift, the distribution of the covariates (features) changes between training and test time, i.e. $P_{tr}(x) \neq P_{tst}(x)$ while leaving the conditional distribution unchanged $P_{tr}(y|x) = P_{tst}(y|x)$ [37], [2].

**Motivation** Real-world data are virtually infinite as they are not upper bounded by the amount of compute, time, space and resources available. This makes reliance on learning systems difficult in environments where stakes are high because of risk on human life, ethical impact on society and required adherence to governance and regulatory compliance. Some significant examples are discussed. In healthcare, varying blood and behavioral markers make heart attack and Alzheimer's disease detection difficult [17]. In criminal justice systems, change in features for recidivism causes, complicate criminal behavior assessment and prediction [32]. Neuropsychologists face unreliability in brain-computer interfacing as features differ not only due to uncontrollable changes in subject's attention but also as to differing electrode placements [15]. Behavioral therapists experience similar problem in Emotion recognition as flux is observed in human expression under multi modal arrangements [14]. Voice biometrics bear speaker recognition complication as speech features vary due temporal, emotional or recording environment changes [21]. Businesses and financial institutions face similar issues in credit risk analysis and spam detection [41]. Thus these motivate the case for minimizing shift in covariates, eventually increasing trust in quality of the decisions made by these learning systems and their suitability for real-world environment.

**Dataset fragmentation and covariate shift** Of particular interest is the setting where the non-colocation of training data in time will induce covariate shift because we may now have several training mini-sets, each with own empirical distribution. The problem lies in cross-validation not providing a reliable estimate of the out-of-sample risk. We are therefore placed within supervised, parametric learning. Moreno-Torres et al. [24] were the first to identify that during *k*-fold cross-validation, each individual fold distribution being slightly unique, produced a different result than hold-out cross validation. Their solutions consisted in designing optimally-balanced cross-validation strategies. That observation served the goal of working around class imbalance, and, to the best of our knowledge, covariate shift induced by temporal dislocation of training data has not been formally studied. Formally, we study the experimental settings where a dataset is fragmented for cross-validation into a number of *batches* or *folds*, i.e. $P_{batch_i}(x) \neq P_{tst}(x)$. We call this *fragmentation-induced-covariate shift (FIcs)*.

**Leveraging divergences between distributions** Sugiyama and many others [37], [7], [10] describe covariate shift to be specified by the density ratio $w(x) = \frac{dP}{dQ}(x)$ that they refer to as *importance weighting*. That quantity is a ratio of distributions, and may not be easy to compute, and impossible for implicit distributions. Usually, the line of work from the prior art on covariate shift proposes approximations to estimate this quantity from data when a simple interval-wise arithmetic division between histograms was not forthcoming. We propose a new approximation to estimate the importance. The estimation is formalized by the $f$-divergence, given $P$ and $Q$ be two probability distributions over a space $\Omega$ such that $P$ is continuous with respect to $Q$. For a convex function $f$ such that $f(1) = 0$, the $f$-divergence of $P$ from $Q$ is defined as $D_f(P||Q) = E_Q f(\frac{dP}{dQ})$ with $\frac{dP}{dQ}$ interpreted as the Radon-Nikodym derivative [34].

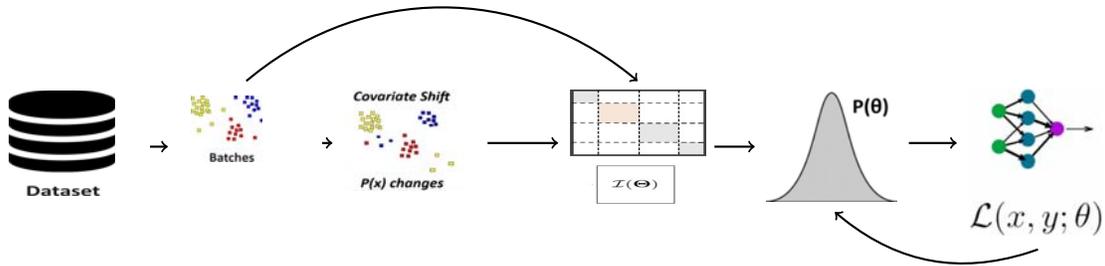

Fig. 1: workflow of FIcsR. We build a global prior to regularise L.

The relevant choice of $f$ include $f(x) = x \log x$, which yields the Kullback–Leibler (KL) divergence [18]. This provides a metric on the space of probability distributions in order to measure the amount of shift between the distribution of different batches and the validation set.

The usual mean-field formulation of the KL-divergence uses a Gaussian distribution as the variational prior $Q(\vartheta)$, parametrised with the covariance matrix of the network parameters. This offers an added benefit, that its computation does not require access to examples from both distributions, in contrast to the maximum-mean discrepancy (MMD) measure being used in importance-weighting methods.

The term however involves the Hessian of the parameters, and its high-dimensionality for usual over-parametrized models puts up an intractability barrier to its computation. Recent inference literature [28], [3] suggests an approximation of the second-order derivative of the KL-divergence by the Fisher Information Matrix (FIM). The latter quantity that can be derived using the variance and expected value of the function of interest [27], here, network parameters. Penalising a function of the FIM between a batch and the validation set or among folds therefore serves as an antidote to fragmentation-induced covariate shift. A high-level overview of our method is given in figure 1.

**Contributions**
1) We study and alleviate for covariate distribution shift induced when fragments of data are separately cross-validated.
2) We present a computationally tractable method for estimating a function of network parameters when trained at one time on a fragment of the entire training data. Additionally, our method allows memory cost to be linear in the size of a sub-dataset.
3) We introduce an information-theoretic penalty containing parameter information for all previous batches to initialize parameter estimates for the current fragment of training data.

## II. DEVELOPMENT

**Notational setup and definitions** Let $x \in X$ be a training example drawn from a complete training dataset $X$. $X$ is further split into $k$ batches $\{B_1 \cdots B\}_k$. $X$ follows a training data distribution $P(X)$, and there exist similar per-batch parameter distributions, i.e. $P(\vartheta_{B_i})$ denotes the distribution of network parameters for batch $B_i$. We may often use $P(\vartheta)$ for ease of readability for any general distribution. $\hat{f}(\vartheta)$ represents a fit to training data (i.e. a learner or trained model), regardless of batching haivng been performed.

**Accumulating parameter priors** If there were no batching, we would have had a single training set distribution and the validation set distribution. We aim to reconstruct an approximation of this distribution, in order to neutralize the effects of FIcs.

If we knew the differences between their parameters, an average over them may provide us with that approximation. We require a quantification of the amount of distribution shift between the batch and validation pairs (the test set is inaccessible during training) of distribution, as batches orchestrate in time. Our method will, at an abstract level, provide an incremental average of the parameters by maintaining a parameter approximation of the entire training data. Let $-$ banally represent a covariate shift, then $P(\vartheta_{val}) - P(\vartheta_{B_1})$ is the covariate shift from the first batch, which potentially exists because of sub-sampling bias. The subsequent batch will have its own $P(\vartheta_{val}) - P(\vartheta_{B_2})$ but also an implied $P(\vartheta_{B_1}) - P(\vartheta_{B_2})$. This has an implication in that any subsequent batch begins its parameter estimation using the previous batch's parameters as priors. Now, if $P(\vartheta_{val}) - P(\vartheta_{B_1})$ may be mitigated, then we have a parameter prior almost free of covariate shift. Therefore, we systematically build a memory of these corrections into our priors to provide any subsequent batch a parameter estimation that lies close to te covariate shift -*free* cross-validation baseline.

**Estimating the covariate shift** The implementation of is the KL-divergence as a priori stated. If we assume $P(x; \vartheta)$ as true distribution for given input $X$ with true parameter $\vartheta$, and $Q(x; \hat{\vartheta})$ as arbitrary target distribution with parameter $\hat{\vartheta}$ then we can rewrite $D_{KL}$ as:

$$D_{KL}(P(\cdot; \vartheta) \| Q(\cdot; \hat{\vartheta})) = E_{X \sim P(\cdot; \vartheta)} \left[ \log \frac{Q(X; \hat{\vartheta})}{P(X; \vartheta)} \right] \quad (1)$$

Consider the special case of $Q(x; \hat{\vartheta})$ parameterized by $\hat{\vartheta}$ whereby we want to minimize $D_{KL}$ w.r.t $\hat{\vartheta}$. For this case $D_{KL}$ is at minimum if we have $Q(x; \hat{\vartheta}) = P(x; \hat{\vartheta})$. Thus we get:

$$D(P(\cdot; \vartheta) \| P(\cdot; \hat{\vartheta})) \geq 0 \quad (2)$$

By applying Taylor expansion up to second-order to the log $P(x; \hat{\vartheta})$ for true parameter $\vartheta$ we have:

$$\log P(X; \hat{\vartheta}) = \log P(X; \vartheta) + (\hat{\vartheta} - \vartheta) \frac{\partial \log P(X; \vartheta)}{\partial \vartheta} - \frac{1}{2}(\hat{\vartheta} - \vartheta)^2 \frac{\partial^2 \log P(X; \vartheta)}{\partial \vartheta^2} + O((\hat{\vartheta} - \vartheta)^3) \quad (3)$$

where usually $y$ is continuous target variable, and $P(y|x)$ and $Q(y|x)$ an arbitrary and a Gaussian distribution, with means $\mu_P$ and $\mu_Q$ and variances $\sigma_P^2$ and $\sigma_Q^2$.

It is crucial to understand the relation of FIM to $D_{KL}$ ($D_{KL}$).

Reiterating the need for an approximation for the KL-divergence for highly parametrised distributions, we revert to [28], [3] for an approximation of the second-order derivative of the KL-divergence by the Fisher Information Matrix (FIM). The Fisher information $I(\vartheta)$ can be defined as the expected value of the negative hessian of the log-likelihood function (see definition 2.4.1, book page 40, [22], see page 116 Eq. 5.16 [20]).

$$I(\vartheta) = E\left[-\frac{\partial^2 \log P(x | \vartheta)}{\partial \vartheta \partial \vartheta^T}\right] \quad (4)$$

The Cramér-Rao Lower Bound Optimization (CRLBO) [8] states that for any unbiased estimator $\hat{\vartheta}$, the variance-covariance matrix $V(\hat{\vartheta})$ satisfies the inequality property (see page 126 eq 6.22 [20], page 3 theorem 15.4 lecture 15 [9]):

$$\quad (5)$$

The symbol $\succeq$ represents the following matrix inequality for $V(\hat{\vartheta}) - I^{-1}(\vartheta)$ positive and semi-definite.

If we are estimating a Gaussian distribution function $q(\hat{\vartheta})$ around parameter $\vartheta$ mean and variance-covariance matrix $V(\hat{\vartheta})$ in such manner that:

$$Q(\hat{\vartheta}) \sim N(\vartheta, V(\hat{\vartheta})) \quad (6)$$

Here, the CRLBO property gives

$$\sigma^2(\hat{\vartheta}) \geq \frac{1}{nI(\vartheta)} \quad (7)$$

where $I(\vartheta)$ is Fisher information and $n$ is sample size.

By taking expectation w.r.t $X$ in Eq. 3 we have:

$$E_{X \sim P(\cdot; \vartheta)}\left[\log P(X; \hat{\vartheta}) - \log P(X; \vartheta)\right] = (\hat{\vartheta} - \vartheta) E_{X \sim P(\cdot; \vartheta)}\left[\frac{\partial \log P(X; \vartheta)}{\partial \vartheta}\right] \quad (8)$$

$$- \frac{1}{2}(\hat{\vartheta} - \vartheta)^2 I(\vartheta) + O((\hat{\vartheta} - \vartheta)^3)$$

The left-hand in above mentioned equation is $D_{KL}$ i.e $D(P(\cdot; \vartheta) \| P(\cdot; \hat{\vartheta}))$. Since $D_{KL}$ is always non-negative, so the right-hand side must also be non-negative. Thus we get:

$$\frac{\hat{\vartheta} - \vartheta}{2} I(\vartheta) \geq 0 \quad (9)$$

It will hold for any $\hat{\vartheta}$, from this we can conclude that:

$$I(\vartheta) \geq 0 \quad (10)$$

which reinforces that the Fisher information is convex, and does not alter the monotonicity of the network loss, if added to it (13).

With the Gaussianity of $Q$ (mean-field variational inference), we may rewrite $D_{KL}$ in terms of parameters:

$$D_{KL}(P(\vartheta) \| Q(\hat{\vartheta})) \sim \int P(\vartheta) \log \frac{P(\vartheta)}{N(\vartheta, V(\hat{\vartheta}))} d\vartheta \quad (11)$$

With the help of CRLB we can replace $V(\hat{\vartheta})$ with $I^{-1}(\vartheta)$ as $V(\hat{\vartheta}) \succeq I^{-1}(\vartheta)$, we get:

$$D_{KL}(P(\vartheta) \| Q(\hat{\vartheta})) \int P(\vartheta) \log \frac{P(\vartheta)}{N(\vartheta, I^{-1}(\tilde{\vartheta}))} d\vartheta \quad (12)$$

which is the estimation of relative entropy by using a variance-covariance matrix of estimated parameters with the help of FIM.

Therefore, the averaging of the batch-to-validation shift correction into the priors could happen with each update of the parameters, via the modified loss. Monotonicity guarantees allow this over all values of the function.

$$L(x, y; \vartheta) = - \int d\vartheta \, P(y) \log(P(y|x; \vartheta)) - \lambda \times \int \frac{\partial^2 \log P(X | \vartheta)}{\partial \vartheta \partial \vartheta^T} \quad (13)$$

The proposed penalized loss offers a significant decrease in computation by using the FIM instead of the entire divergence term.

## III. BENCHMARKS AND STATE-OF-THE-ART

Few published works addresses the covariate shift mitigation problem. Most of the work is through weight estimation by statistical optimization approaches [38], [13], [7]. Empirical risk minimization is a standard method in supervised learning but it fails under covariate shift where features training and test distribution differs [33], [39]. To mitigate covariate shift, [35], [23], [37], [45] uses weight adjusting, where they assigns importance weighting (IW) to each training example $w(x) = \frac{p_{tst}(x)}{p_{trn}(x)}$ is described as *importance* which leads to *importance weighted* ERM (IW-ERM): $\min_{f \in F} \frac{1}{n_{tr}} \sum_{j=1}^{n_{trn}} \ell(f(x_j^{trn}), y_j^{trn}) w(x^{trn})$, However IWERM often results in high variance which inflates test risk [35], [36]. [35] replace importance weighting with $(\frac{p_{tst}(x)}{p_{trn}(x)})^\lambda$ where $\lambda \in [0, 1]$ is flattening parameter to reduce the variance for balancing bias-variance trade-off, which described as *exponentially-flattened importance weighted* ERM (EIWERM): $\min_{f \in F} \frac{1}{n_{trn}} \sum_{j=1}^{n_{trn}} \ell(f(x_j^{trn}), y_j^{trn}) w(x_j^{trn})^\lambda$, One of the key idea for importance based methods is how to estimate importance accurately. [38] proposed Unconstrained Least-Squares Importance Fitting (uLSIF) method to estimate importance and adjust to target distribution. [44] proposed *relative importance weighted* ERM (RIWERM) which finds the flattens estimates directly. The relative importance $w_\lambda(x) = \frac{p_{tst}(x)}{(1-\lambda)p_{tst}(x)+\lambda p_{trn}(x)}$. The relative importance $w_\lambda(x)$ can be estimted using relative uLSIF (RuLSIF) [44]. Sugiyama et al. [46] identifies and mitigates limitations in existing methods by proposing a *One-step* approach that seamlessly integrates importance estimation with empirical risk minimization, effectively bounding the test risk. One-step method assumes a hypothesis set $F \subseteq \{f : X \to R\}$ and imposes a boundedness condition $\ell(f(x), y) \leq m$ on the loss function. If the distribution shifts occurs significantly between datasets, then the assumptions $m$ may no longer hold, potentially constraining the model $f$'s flexibility. [24] investigates the problem of induced shift caused by the $k$-fold cross-validation. They use different variants of cross-validation for adapting to induced-covariate-shift. [47], change the test set distribution continuously; we do not benchmark with this due to our batch focused approach. All of these approaches assume datasets' in-memory presence. To the best of our knowledge, none of these works address the problem where data in high-volume settings are shown to different networks and need reconciliation.

## IV. EXPERIMENTAL SETUP

We demonstrate the efficacy of *FIcsR* against multiple baseline settings for fragmentation-induced covariate shift and on the benchmarks for standard covariate shift as a surrogate.

### A. Model architecture and hyper-parameter selection:

We used a five-layer convolutional neural network (CNN) with softmax cross-entropy loss. Our CNN model consists of 2 convolutional layers with pooling and 3 fully connected layers. The model architecture remains consistent image-based benchmarks. The hyper-parameter settings for all image-based datasets remains the same i.e for image datasets we keep hyper-parameters (optimizer = Adam, activation = softmax and epochs = 100). However, for tabular datasets the model architecture differs from image-based but remains the same for all tabular datasets. We use a multi-layer perceptron network for tabular data with a hidden layer with 4 neurons. In tabular settings the hyper-parameter setup is (activation='relu', optimizer ='adam', epochs=1500), the mean accuracy averaged over 100 trials for each experiment. The penalty hyper-parameter $\lambda$ for all datasets remains the same. We calibrate $\lambda$ value with batch/fold setup as results reported in fig 2 . All baselines are implemented in TensorFlow 2.11 and code can be found on the given anonymous link [1]. We reproduce results over the comparing methods as published.

[1] https://anonymous.4open.science/r/FIcsR-D63B

### B. Machine Specification:

We run all of our experiments on RTX 3090 Ti with 24 GB GPU memory and 128 GB system memory. The average runtime for each experiment is 30 minutes for the integral training set and 210 minutes for the fragmented training set. The running time may vary as it depends on the type of datasets and number of batch splits.

### C. Datasets and Evaluation Metric:

*FIcsR* performance evaluated on 16 image-based and 32 binary datasets benchmarks. In three image-based datasets like F-MNIST [42], K-MNIST [4] and MNIST-C [25] covariate shift is induced using Sugiyama et al. One-step approach [46][2] and in five binary datasets like australian , breast-cancer, diabetes, heart and sonar from keel repository [1] covariate shift is induced following the similar way of Sugiyama et al. [46] which he adpated from Cortes et al. [7]. All other dataset like (MNIST [19], EMNIST [6], QMNIST [43], KannadaMNIST [29], CIFAR-10 and CIFAR-100 [5], SVHN [26], Caltech101 [11], Tiny-ImageNet [16], STL-10 [16] , P-MNIST [25] , CIFAR10-C and CIFAR100-C [12]), and 27 benchmarking binary datasets [1] from KEEL repository are used as per their publsihed setting.

Accuracy is used as an evaluation metric to measure and benchmark the performance of *FIcsR*.

### D. Experimental design:

We design and execute a set of five major experiments, using multiple baselines for cross-dimentional comparison.
The experiments see training data in various guises:
- **integral** i.e. unfragmented training set.
- **fragmented** training set, in either cross-validation setting.

Across experimental results, we frequently refer to two cross-validation settings:
Standard cross-validation (**st-CV**) with our penalty not having been applied.
Penalized cross-validation (**FIcsR** to save on column width). Penalized as in eq 13. Note that these may apply indiscriminate to dataset fragmentation, and cross-validation techniques (hold-out or $k$-fold). The interplay of the source of covariate shift and the orchestration of training data give us three baselines for our experiments:
- **BL1:** st-CV on integral training sets.
- **BL2:** st-CV on fragmented training set.
- **BL3:** CV in $k$-fold setting.

The following experiments evaluate distinct aspects of the problem:
**E1:** Evidencing covariate shift induction due to batching. Does batches induce FIcs? We execute *standard* cross-validation (st-CV) on the integral training sets which we consider as baseline (**BL1**). We then perform st-CV on the fragmented training set. Each training set was fragmented with the ratio (5%, 10%, 20%, 25%, 50%), which results in (20, 10, 5, 4, 2) number of batches of the training set. The batches are represented as ($B_1$, $B_2$, $B_3$, $B_{\frac{n}{2}}$ & $B_{n-1}$ and $B_n$) in Table III, IV, VII and VIII, where $n$ is the number of batches mentioned in former sentence. The validation set remains unchanged as per the standard. We compare st-CV on fragmented training set with (**BL1**). To validate (**BL1**), we perform st-CV on 16 image-based and 32 binary datasets. The results are presented in Table III for batches (20, 10, 2). More detailed batchwise results are presented in Table VII.

[2] Each training image $I_i$ is rotated by angle $\vartheta_i$, with $\vartheta_i/180°$ drawn from distribution Beta$(a, b)$. For test images $J_i$, the rotation angle $\phi_i$ is drawn from Beta$(b, a) = (2, 4), (2, 5),$ and $(2, 6)$.

**E2:** Mitigation of FIcs using FIcsR.
Does FIcsR correct the FIcs? We execute our proposed method FIcsR on the integral training set and fragmented training set. We execute FIcsR on no-covariate-shift and standard-covariate-shift datasets to observe that FIcsR can correct standard-covariate-shift. We executed FIcsR on the fragmented training set and compared it with st-CV on the fragmented set baseline (**BL2**). The results are reported in Table IV for batches (20,10,2). To ensure better performance of *FIcsR* we compare the mean accuracy over all batches $\mu_1$ of Table III and $\mu_2$ of Table IV. We report accuracy for each single batch as well in all experimental settings to verify *FIcsR* performance. We may consider *FIcsR* with the whole dataset as a secondary baseline for our *FIcsR* batchwise method.

**E3:** Evidencing the aggravation of FIcs in $k$-fold CV.
Does $k$-fold CV aggravate FIcs? We fragmented no-covariate-shift unfragmented and standard-covariate-shift datasets into number of folds like (2, 5, 10). To evaluate (**BL3**), we executed CV foldwise and compared the results against (**BL1**) to validate that accuracy falls as the data is fragmented into folds. The results are reported in Table IX.

**E4:** Correcting FIcs in folds setting.
Does FIcsR correct FIcs in the folds setting? To observe the performance of FIcsR we executed FIcsR in folds setting with folds $k$ in range (2,5, 10). We compare FIcsR with CV in $k$-fold setting with baseline (**BL3**). Foldwise results are presented in (Tables V, VI, IX and X).

[24] hypothesized that covariate shift is induced due to k-fold setup for cross-validation. In the foldwise setting, the test set is absent as the reference for measuring the fragmentation-induced covariate shift, while in batchwise settings we have a reference test set.

**E5:** Positioning FIcsR with State-of-the-art.
We benchmark FIcsR performance against state-of-art One-step, EIWERM and RuLSIF methods (Sec III) on same datasets as used by these methods. The results are reported in I and Tables II. Standard deviation over multiple runs parentheses following the value of the mean.

## V. RESULTS AND DISCUSSION

### A. Evidencing covariate shift induction due to batching

**No-covariate-shift datasets:** As can be observed from Table III, fragmenting data results in decreasing both, average and batchwise accuracy when compared with st-CV baseline (BL1). This can be argued for the covariate shift induced using fragmentation operation. Results report more than 36% decrease in average accuracy over 20, 10, and 2 batch fragmentation.

**Standard-covariate-shift datasets:** The aforementioned pattern is observed to be consistent across standard-covariate-shift datasets also, yet comparatively following a much steeper drop in average accuracy. Results from Table III report more than 60% decrease in average accuracy over 20, 10 and 2 batch fragmentation against st-CV baseline (BL1). This sharp drop can be a compounding effect brought about by standard-covariate-shift deterioration as additional shift is induced via fragmentation operation.

**Correlation with batch frequency:** An interesting observation from Figure 3a highlights a positive correlation between the decrease in average accuracy and batch fragmentation frequency. These results tabulated in Table III show that a high drop in accuracy of 52.1% over 20 batches moderates to 43.70% over 10 batches and dilutes to 36.30% over 2 batches, on no-covariate-shift datasets. This can be argued for the large data support resulting from few batches thereby offsetting the induced shift. The pattern is observed to be consistent across all the reported datasets.

In Table III and IV (discussed below), results are reported in alternate batches for the benefit of space. An extended version of these tables can be found in Table VII and VIII, respectively.

### B. Mitigation of FIcs using FIcsR

**FIcs mitigation:** *FIcsR*, improving upon the baseline (BL2), successfully mitigates FIcs on no-covariate-shift datasets. As shown in Table IV, column $\Delta_3$, this effect of remediation results in an increase in average accuracy by more than 10% under 20, 10, and 2 batch fragmentation settings. Furthermore, the observed improvement is sustainable across datasets. This performance gain can be attributed to *FIcsR*'s capability of absorbing and retaining maximum information from the batch sequence while regularizing the model, leading to effective remediation.

**Double mitigation:** *FIcsR* successfully improves upon the baseline (BL2) under more challenging conditions where the presence of standard shift is aggravated via fragmentation operation on standard-covariate-shift datasets. Table IV reports a performance increase of more than 25% under 20, 10 and 2 batch fragmentation. This surge in performance is a compound effect as *FIcsR*' mitigates both, standard and fragmented induced covariate shift, hence referred as "double mitigation".

**Consistency**: Running a cross batch comparison between a "with" and "remediated" shift batch, shows significant performance improvement in the latter. The results are observed to be consistent over any comparison, provided the batch pair belongs to the same position in the fragmentation sequence. Moreover, this improvement is consistent under multiple fragmentation settings and across datasets. For example, under 20 batch fragmentation on F-MNIST, the batch pair $B_n$ reports an accuracy of 70.6% "with shift" (Table III) and 81.9% "remediated shift" (Table IV), clearly establishing the effect of remediation, which resulted in performance improvement of 16%. This consistency can be attributed to *FIcsR*'s capability of regularizing model parameters from current data and sustaining its performance across batch sequence.

TABLE I: Benchmarking with SOTA on image datasets ($\Delta_1$ % = FIcsR - SOTA)

| Dataset | Shift Level (a, b) | ERM | EIWERM | One-step | FIcsR | $\Delta_1$% |
|---|---|---|---|---|---|---|
| F-MNIST | (2, 4) | 64.6(0.17) | 71.3(0.06) | 74.5(0.08) | 78.3(0.14) | ↑ 5.10% |
| | (2, 5) | 54.5(0.54) | 57.9(0.29) | 55.6(0.20) | 57.2(0.31) | ↑ 2.87% |
| | (2, 6) | 36.3(0.34) | 42.5(0.55) | 44.8(0.25) | 45.4(0.20) | ↑ 1.33% |
| K-MNIST | (2, 4) | 67.1(0.18) | 69.7(0.24) | 68.8(0.12) | 72.2(0.44) | ↑ 4.94% |
| | (2, 5) | 55.0(0.26) | 52.2(0.19) | 59.5(0.16) | 61.6(0.09) | ↑ 3.52% |
| | (2, 6) | 39.2(0.30) | 38.4(0.93) | 43.1(0.55) | 43.8(0.35) | ↑ 1.62% |
| MNIST-C | (2, 4) | 63.6(0.91) | 80.5(0.08) | 85.2(0.17) | 86.3(0.28) | ↑ 1.29% |
| | (2, 5) | 43.8(0.18) | 60.4(0.47) | 78.4(0.32) | 79.3(0.67) | ↑ 1.14% |
| | (2, 6) | 33.3(0.49) | 53.8(0.13) | 64.2(0.67) | 64.9(0.37) | ↑ 1.09% |

TABLE II: Benchmarking with SOTA on tabular datasets. Mean accuracy with *Wilcoxon signed-rank test*[40] at significance level 5% across various datasets with induced covariate shift ($\Delta_2$ % = FIcsR - SOTA)

| Dataset | ERM | uLSIF | RuLSIF | One-step | FIcsR | $\Delta_2$% |
|---|---|---|---|---|---|---|
| australian | 67.9 (16.8) | 69.3 (16.3) | 69.6 (15.1) | 74.4 (12.7) | **75.7 (5.86)** | ↑ 1.74% |
| breast cancer | 78.3 (13.4) | 79.9 (12.4) | 78.6 (12.9) | 77.4 (10.1) | 73.6 (4.62) | ↓ 4.90% |
| diabetes | 54.2 (8.88) | 57.5 (7.66) | 55.7 (8.63) | 62.9 (6.36) | **64.3 (12.6)** | ↑ 2.22% |
| heart | 65.28 (9.91) | 64.1 (11.4) | 63.2 (11.7) | 74.3 (10.9) | **78.1 (5.90)** | ↑ 5.11% |
| sonar | 61.9 (12.9) | 64.6 (13.2) | 63.7 (13.5) | 67.6 (12.4) | **70.4 (5.91)** | ↑ 4.14% |

TABLE III: st-CV batch-wise accuracy

| Dataset | Baseline st-CV | Batchwise accuracy $B_1$ | $B_2$ | $B_{\frac{n}{2}}$ | $B_{n-1}$ | $B_n$ | Mean $\mu_1$ | var $\sigma_1^2$ | $\Delta$% st-CV $- \mu_1$ |
|---|---|---|---|---|---|---|---|---|---|
| \multicolumn{10}{c}{Training data = 5% , Number_of_Batches = 20} | | | | | | | | | |
| MNIST | 94.8 | 89.3 | 87.9 | 89.9 | 88.9 | 88.8 | 88.7 | 0.49 | ↓ 6.43 |
| EMNIST | 83.1 | 73.7 | 74.2 | 72.5 | 74.0 | 70.6 | 72.9 | 1.94 | ↓ 12.2 |
| CIFAR-10 | 71.5 | 49.0 | 50.3 | 50.7 | 51.2 | 54.5 | 49.9 | 9.67 | ↓ 30.2 |
| CIFAR-100 | 38.2 | 16.5 | 18.1 | 19.3 | 20.4 | 22.7 | 18.3 | 9.17 | ↓ 52.1 |
| P-MNIST | 95.1 | 86.1 | 88.9 | 88.7 | 87.2 | 88.3 | 88.4 | 1.41 | ↓ 7.04 |
| QMNIST | 75.4 | 63.4 | 62.2 | 66.7 | 58.3 | 63.4 | 63.4 | 5.59 | ↓ 15.9 |
| CIFAR10-C | 63.9 | 20.1 | 16.3 | 16.1 | 14.9 | 10.2 | 16.2 | 10.4 | ↓ 74.6 |
| CIFAR100-C | 28.8 | 16.3 | 19.1 | 19.7 | 21.6 | 22.3 | 18.4 | 14.2 | ↓ 36.1 |
| \multicolumn{10}{c}{Training data = 10% , Number_of_Batches = 10} | | | | | | | | | |
| MNIST | 94.8 | 91.7 | 91.1 | 90.2 | 89.3 | 91.5 | 91.1 | 1.06 | ↓ 3.90 |
| EMNIST | 83.1 | 69.9 | 75.2 | 73.4 | 74.4 | 71.7 | 73.7 | 9.22 | ↓ 11.3 |
| CIFAR-10 | 71.5 | 52.8 | 52.9 | 53.7 | 54.5 | 54.1 | 52.6 | 4.87 | ↓ 26.4 |
| CIFAR-100 | 38.2 | 20.3 | 22.2 | 23.3 | 24.7 | 21.2 | 21.5 | 5.51 | ↓ 43.7 |
| P-MNIST | 95.1 | 92.8 | 91.8 | 91.2 | 91.4 | 91.3 | 91.5 | 0.62 | ↓ 3.78 |
| QMNIST | 75.4 | 63.6 | 64.1 | 64.9 | 65.2 | 64.5 | 64.0 | 1.15 | ↓ 15.1 |
| CIFAR10- C | 63.9 | 17.6 | 18.4 | 12.2 | 17.4 | 12.9 | 22.6 | 16.8 | ↓ 64.6 |
| CIFAR100-C | 28.8 | 21.9 | 25.2 | 26.6 | 27.1 | 20.5 | 22.8 | 16.3 | ↓ 20.8 |
| \multicolumn{10}{c}{Training data = 50% , Number_of_Batches = 2} | | | | | | | | | |
| MNIST | 94.8 | 93.3 | 93.7 | ˇ | ˇ | ˇ | 93.5 | 0.08 | ↓ 1.37 |
| EMNIST | 83.1 | 80.0 | 79.7 | ˇ | ˇ | ˇ | 79.8 | 0.04 | ↓ 3.97 |
| CIFAR-10 | 71.5 | 56.2 | 60.1 | ˇ | ˇ | ˇ | 58.1 | 3.81 | ↓ 18.7 |
| CIFAR-100 | 38.2 | 23.2 | 25.4 | ˇ | ˇ | ˇ | 24.3 | 1.21 | ↓ 36.3 |
| P-MNIST | 95.1 | 93.3 | 93.7 | ˇ | ˇ | ˇ | 93.5 | 0.08 | ↓ 1.68 |
| QMNIST | 75.4 | 73.6 | 73.3 | ˇ | ˇ | ˇ | 73.5 | 0.04 | ↓ 2.51 |
| CIFAR10- C | 63.9 | 49.0 | 41.3 | ˇ | ˇ | ˇ | 45.1 | 29.6 | ↓ 29.1 |
| CIFAR100- C | 28.8 | 25.3 | 27.5 | ˇ | ˇ | ˇ | 26.4 | 1.21 | ↓ 8.33 |

*C. Evidencing the aggravation of FIcs in k-fold CV*

Table V presents st-CV results in folds setting. We consider st-CV (**BL1**) baseline for folds settings to validate our hypothesis that shift is induced due to the dats fragmentation. Table V & IX shows that as a dataset is fragmented folds, accuracy falls as compared to (**BL1**), which indicates that shift is induced. We report average accuracy for each fold setting and single fold also. $\mu_3, \mu_4, \mu_5$ shows the average accuracy of st-CV for 2 folds, 5 folds, and 10 folds settings.

*D. Correcting FIcs in folds setting*

**Tabular datasets:** The results for (k = 2, 5, 10) folds settings for tabular datasets are reported in Tables (VI & X).

In table VI, $\Delta_5 = (\mu_3 - \mu_6)$, $\Delta_6 = (\mu_4 - \mu_7)$, and $\Delta_5 = (\mu_5 - \mu_8)$ show difference in average accuracy in folds setting. We improve by a maximum of 28.3% in all *k*-fold settings.

**Image datasets:** *FIcsR* results for 13 image-based datasets are reported in Table X. We may consider st-cv foldwise as the secondary baseline for *FIcsR* foldwise. *FIcsR* accuracy improvement is shown as $\Delta_4 = (\mu_9 - \mu_{10})$ in Table X. We improve by a maximum of 43.2% in all *k*-fold settings.

*E. Comparison with State-of-the-art*

**Image datasets** Our method *FIcsR*, significantly outperforms state-of-the-art namely EIWERM and One-step. As can be observed from Table I , column $\Delta_1$, *FIcsR* reports a

TABLE IV: *FIcsR* Batchwise

| Dataset | *FIcsR* | Batchwise accuracy | | | | | | Mean | var | $\Delta_3 = \mu_2 - \mu_1$ |
|---|---|---|---|---|---|---|---|---|---|---|
| | | $B_1$ | $B_2$ | $B_3$ | $B_{\frac{n}{2}}$ | $B_{n-1}$ | $B_n$ | $\mu_2$ | $\sigma_2^2$ | $\Delta_3(\%)$ |
| Training data = 5% , Number_of_Batches = 20 | | | | | | | | | | |
| MNIST | 97.9 | 90.7 | 90.6 | 91 | 91.7 | 91.4 | 91.8 | 91.2 | 0.09 | ↑ 2.81 |
| EMNIST | 88.4 | 81.5 | 81.7 | 81.2 | 81.4 | 81.5 | 81.9 | 81.5 | 0.06 | ↑ 11.7 |
| CIFAR-10 | 87.7 | 50.9 | 51.4 | 52.2 | 48.9 | 50.3 | 57.4 | 51.8 | 7.18 | ↑ 3.81 |
| CIFAR-100 | 58.7 | 23.9 | 18.2 | 18.5 | 17.8 | 23.9 | 18.3 | 20.1 | 7.26 | ↑ 9.83 |
| P-MNIST | 97.6 | 91.1 | 89.8 | 90.2 | 90.4 | 91.5 | 90.7 | 90.3 | 0.26 | ↑ 2.14 |
| QMNIST | 89.2 | 68.5 | 69.4 | 67.2 | 68 | 67.8 | 66.9 | 68.4 | 0.63 | ↑ 7.88 |
| CIFAR10-C | 73.3 | 46.4 | 54.3 | 57.8 | 61.1 | 61.5 | 61.8 | 57.2 | 30.1 | ↑ 253 |
| CIFAR100-C | 39.4 | 11.9 | 17.2 | 18.5 | 21.3 | 22.1 | 24.9 | 19.3 | 17.1 | ↑ 4.89 |
| Training data = 10% , Number_of_Batches = 10 | | | | | | | | | | |
| MNIST | 97.9 | 91.9 | 91.7 | 91.2 | 91.8 | 91.3 | 91.8 | 91.7 | 0.08 | ↑ 0.65 |
| EMNIST | 88.4 | 79.5 | 82.4 | 81.6 | 79.5 | 82.3 | 81.9 | 81.2 | 1.21 | ↑ 10.1 |
| CIFAR-10 | 87.7 | 52.1 | 53.1 | 48.5 | 59.3 | 52.5 | 55.7 | 53.5 | 11.1 | ↑ 1.71 |
| CIFAR-100 | 58.7 | 27.2 | 25.8 | 20.4 | 17.0 | 21.9 | 22.8 | 22.5 | 11.3 | ↑ 4.65 |
| P-MNIST | 97.6 | 91.6 | 91.9 | 91.3 | 91.6 | 90.1 | 91.2 | 91.5 | 0.31 | 0.00 |
| QMNIST | 89.2 | 71.4 | 70.4 | 71.7 | 70.7 | 70.5 | 70.9 | 70.9 | 0.71 | ↑ 10.7 |
| CIFAR10-C | 73.3 | 52.7 | 59.9 | 61.9 | 64.4 | 66.1 | 65.7 | 61.7 | 21.1 | ↑ 173 |
| CIFAR100-C | 39.4 | 16.2 | 22.1 | 24.8 | 27.2 | 26.8 | 27.3 | 21.1 | 15.6 | ↓ 7.45 |
| Training data = 50% , Number_of_Batches = 2 | | | | | | | | | | |
| MNIST | 97.9 | 95.9 | 96.1 | ˇ | ˇ | ˇ | ˇ | 96.0 | 0.02 | ↑ 2.67 |
| EMNIST | 88.4 | 84.2 | 84.4 | ˇ | ˇ | ˇ | ˇ | 84.3 | 0.02 | ↑ 5.63 |
| CIFAR-10 | 87.7 | 76.3 | 80.6 | ˇ | ˇ | ˇ | ˇ | 78.4 | 4.62 | ↑ 34.9 |
| CIFAR-100 | 58.7 | 39.8 | 39.9 | ˇ | ˇ | ˇ | ˇ | 39.85 | .002 | ↑ 63.9 |
| P-MNIST | 97.6 | 95.7 | 96.1 | ˇ | ˇ | ˇ | ˇ | 95.9 | 0.08 | ↑ 2.56 |
| QMNIST | 89.2 | 79.3 | 80.4 | ˇ | ˇ | ˇ | ˇ | 79.8 | 0.61 | ↑ 8.57 |
| CIFAR10-C | 73.3 | 65.5 | 68.6 | ˇ | ˇ | ˇ | ˇ | 67.1 | 2.40 | ↑ 4.35 |
| CIFAR100-C | 39.4 | 31.2 | 34.8 | ˇ | ˇ | ˇ | ˇ | 33.0 | 3.24 | ↑ 25.0 |

TABLE V: st-CV foldwise

| Dataset | Baseline | k=2 | | | k=5 | | | | | | k=10 | | | | | |
|---|---|---|---|---|---|---|---|---|---|---|---|---|---|---|---|---|
| | st-CV | $k_1$ | $k_2$ | $\mu_3$ | $k_1$ | $k_2$ | $k_3$ | $k_4$ | $k_5$ | $\mu_4$ | $k_1$ | $k_2$ | $k_{\frac{n}{2}}$ | $k_{n-1}$ | $k_n$ | $\mu_5$ |
| Appendicitis | 98.1 | 97.6 | 97.6 | 97.6 | 97.8 | 98.8 | 96.7 | 97.8 | 98.5 | 97.9 | 96.2 | 99.2 | 97.1 | 97.8 | 98.5 | 98.0 |
| Lymphography | 85.5 | 81.4 | 83.2 | 82.3 | 84.1 | 84.1 | 87.6 | 84.7 | 86.9 | 85.5 | 85.5 | 84.1 | 88.4 | 79.7 | 89.8 | 85.3 |
| Banana | 77.9 | 70.9 | 75.8 | 73.4 | 71.4 | 70.4 | 76.6 | 72.5 | 75.8 | 73.3 | 70.7 | 70.0 | 72.1 | 73.0 | 70.3 | 71.7 |
| Bands | 81.5 | 68.8 | 71.1 | .700 | 71.2 | 70.3 | 71.2 | 63.8 | 74.1 | 70.1 | 68.5 | 64.8 | 70.3 | 57.4 | 77.7 | 74.1 |
| LiverDisorders | 65.5 | 59.4 | 57.3 | 58.3 | 62.1 | 61.4 | 57.8 | 56.1 | 66.6 | 60.8 | 65.5 | 58.6 | 65.5 | 62.1 | 64.2 | 61.5 |
| Bupa | 54.5 | 64.2 | 51.8 | 58.1 | 63.6 | 54.5 | 81.8 | 36.3 | 63.6 | 60.0 | 33.3 | 33.3 | 80.0 | 60.0 | 40.0 | 61.3 |
| Chess | 98.4 | 93.3 | 93.9 | 93.6 | 95.0 | 96.4 | 97.8 | 98.1 | 97.1 | 96.8 | 96.5 | 98.7 | 98.7 | 97.8 | 98.4 | 97.9 |
| CrX | 84.7 | 75.3 | 73.1 | 74.2 | 79.7 | 79.7 | 73.9 | 81.8 | 84.1 | 79.8 | 79.7 | 79.7 | 76.8 | 84.1 | 81.1 | 82.4 |
| GermmanCredit | 70.5 | 69.8 | 68.6 | 69.2 | 75.5 | 70.0 | 65.0 | 73.5 | 72.5 | 71.3 | 72.0 | 77.0 | 65.0 | 78.0 | 70.0 | 71.7 |
| Haberman | 69.4 | 73.8 | 70.5 | 72.2 | 70.9 | 77.1 | 73.7 | 75.4 | 72.1 | 73.8 | 77.4 | 77.4 | 58.1 | 73.3 | 76.6 | 74.1 |
| Statlog(Heart) | 83.3 | 59.2 | 51.8 | 55.5 | 64.8 | 64.8 | 64.8 | 57.4 | 62.9 | 62.9 | 66.6 | 62.9 | 66.6 | 51.8 | 66.6 | 66.6 |
| Heptatis | 74.2 | 73.1 | 70.1 | 71.6 | 80.6 | 70.9 | 74.2 | 74.2 | 77.4 | 75.4 | 68.7 | 68.7 | 80.0 | 66.6 | 80.0 | 76.7 |
| Housevote | 90.8 | 88.5 | 86.2 | 87.4 | 90.8 | 89.6 | 91.9 | 85.1 | 83.9 | 88.2 | 90.9 | 95.4 | 95.3 | 90.6 | 79.1 | 88.1 |
| Ionosphere | 84.5 | 61.4 | 70.8 | 66.1 | 80.3 | 64.2 | 80.0 | 80.0 | 84.2 | 77.7 | 82.8 | 57.1 | 71.4 | 91.4 | 82.8 | 78.9 |
| Mammographic | 79.7 | 76.7 | 75.0 | 75.8 | 78.2 | 74.4 | 76.0 | 77.1 | 75.5 | 76.2 | 77.1 | 77.1 | 79.2 | 79.2 | 78.1 | 76.6 |
| Monk-2 | 94.6 | 65.4 | 70.8 | 68.1 | 65.1 | 75.6 | 59.4 | 83.7 | 71.1 | 71.1 | 69.6 | 78.5 | 57.1 | 89.1 | 72.7 | 72.5 |
| Mushroom | 99.1 | 100 | 96.7 | 98.3 | 100 | 100 | 100 | 100 | 100 | 100 | 100 | 100 | 100 | 100 | 100 | 100 |
| Phoneme | 80.6 | 80.7 | 78.8 | 79.7 | 82.7 | 80.5 | 80.6 | 80.6 | 80.3 | 80.9 | 82.9 | 79.8 | 79.1 | 81.8 | 81.5 | 80.6 |
| Pima | 74.0 | 71.4 | 69.5 | 70.4 | 72.1 | 76.6 | 73.3 | 78.4 | 72.5 | 74.6 | 76.6 | 84.4 | 68.8 | 76.6 | 71.1 | 74.9 |
| Saheart | 73.1 | 75.7 | 65.8 | 70.7 | 77.4 | 70.9 | 80.4 | 68.4 | 60.8 | 71.6 | 78.7 | 69.5 | 78.2 | 65.2 | 65.2 | 70.5 |
| Thyroid | 73.8 | 80.7 | 75 | 77.8 | 88.1 | 85.7 | 80.9 | 82.9 | 73.1 | 82.2 | 80.9 | 85.7 | 85.7 | 76.1 | 65.0 | 81.6 |
| Spambase | 94.1 | 92.3 | 93.3 | 92.8 | 93.5 | 93.2 | 91.7 | 93.2 | 93.4 | 93.1 | 93.6 | 93.2 | 95.0 | 94.5 | 93.6 | 93.2 |
| SPECTHeart | 74.1 | 68.6 | 65.4 | 67.0 | 70.4 | 72.2 | 64.2 | 75.5 | 58.5 | 68.2 | 74.1 | 70.3 | 66.6 | 80.7 | 57.6 | 68.1 |
| Tic-Tac-Toe | 73.9 | 59.1 | 64.1 | 61.5 | 63.5 | 60.4 | 72.9 | 63.4 | 61.7 | 64.4 | 68.7 | 65.6 | 73.9 | 55.2 | 65.2 | 64.7 |
| Titanic | 73.4 | 78.0 | 77.0 | 77.5 | 72.5 | 78.8 | 78.8 | 77.2 | 77.7 | 77.1 | 77.7 | 78.6 | 75.9 | 76.3 | 76.8 | 76.8 |
| Wdbc | 97.3 | 96.1 | 95.1 | 95.6 | 95.6 | 98.2 | 98.2 | 97.3 | 92.9 | 96.4 | 96.4 | 98.2 | 98.2 | 96.4 | 89.2 | 96.4 |
| Wisconsin | 96.4 | 95.7 | 95.1 | 95.4 | 95.7 | 97.8 | 95.7 | 95.7 | 93.5 | 95.7 | 95.7 | 98.5 | 97.1 | 94.3 | 94.2 | 95.7 |

TABLE VI: *FIcsR* foldwise

| Dataset | Baseline | k = 2 | | | k = 5 | | | | | | k = 10 | | | | | | Δ | | |
|---|---|---|---|---|---|---|---|---|---|---|---|---|---|---|---|---|---|---|---|
| | | $k_1$ | $k_2$ | $\mu_6$ | $k_1$ | $k_2$ | $k_3$ | $k_4$ | $k_5$ | $\mu_7$ | $k_1$ | $k_2$ | $k_{\frac{n}{2}}$ | $k_n-1$ | $k_n$ | $\mu_8$ | $\Delta_5$ | $\Delta_6$ | $\Delta_7$ |
| Appendicitis | 99.6 | 99.1 | 98.8 | 98.9 | 98.5 | 100 | 98.2 | 98.2 | 98.5 | 98.6 | 97.1 | 100 | 98.5 | 100 | 100 | 99.2 | ↑1.30 | ↑0.70 | ↑1.20 |
| Lymphography | 91.3 | 86.6 | 84.1 | 85.3 | 85.5 | 85.5 | 87.6 | 84.7 | 88.4 | 86.3 | 86.9 | 85.5 | 86.9 | 73.9 | 92.7 | 86.2 | ↑3.00 | ↑0.80 | ↑0.90 |
| Banana | 76.3 | 73.1 | 82.1 | 77.6 | 75.2 | 70.9 | 77.5 | 72.1 | 76.6 | 74.5 | 82.6 | 56.0 | 76.6 | 70.5 | 81.5 | 74.1 | ↑4.20 | ↑1.20 | ↑2.40 |
| Bands | 76.8 | 77.4 | 68.8 | 73.1 | 77.7 | 74.1 | 67.5 | 72.2 | 78.8 | 74.1 | 72.2 | 66.6 | 77.7 | 64.8 | 74.1 | 71.6 | ↑3.10 | ↑4.00 | ↓2.50 |
| LiverDisorders | 63.7 | 58.7 | 60.1 | 59.4 | 46.5 | 61.4 | 59.6 | 52.6 | 64.9 | 57.1 | 51.7 | 58.6 | 58.6 | 46.4 | 60.7 | 59.4 | ↑1.10 | ↓3.70 | ↓2.10 |
| Bupa | 72.7 | 57.1 | 59.2 | 58.2 | 72.7 | 63.6 | 54.5 | 36.3 | 63.6 | 58.2 | 66.6 | 33.3 | 0.00 | 20.0 | 40.0 | 55.0 | ↑0.00 | ↓1.8 | ↓6.30 |
| Chess | 98.7 | 98.6 | 98.4 | 98.5 | 98.1 | 99.3 | 99.2 | 98.4 | 97.1 | 98.4 | 98.4 | 98.7 | 99.3 | 99.1 | 98.4 | 99.1 | ↑4.90 | ↑1.6 | ↑1.20 |
| CrX | 84.1 | 84.3 | 87.2 | 85.7 | 86.9 | 86.9 | 84.7 | 90.5 | 83.3 | 86.5 | 86.9 | 91.3 | 82.6 | 86.9 | 86.9 | 86.9 | ↑11.5 | ↑6.7 | ↑4.50 |
| GermmanCredit | 74.0 | 72.2 | 71.6 | 71.9 | 75.5 | 71.0 | 68.0 | 72.5 | 75.0 | 72.4 | 73.0 | 76.0 | 66.0 | 74.0 | 76.0 | 72.8 | ↑2.70 | ↑1.10 | ↑1.10 |
| Haberman | 72.5 | 77.1 | 70.5 | 73.8 | 69.4 | 75.4 | 73.7 | 77.1 | 75.4 | 74.2 | 67.7 | 80.6 | 64.5 | 73.3 | 73.3 | 74.1 | ↑1.60 | ↑0.40 | ↑0.00 |
| Statlog(Heart) | 87.1 | 77.1 | 80.7 | 78.8 | 85.2 | 72.2 | 88.8 | 85.2 | 75.9 | 81.5 | 92.5 | 74.1 | 85.1 | 85.1 | 77.7 | 82.9 | ↑23.3 | ↑18.6 | ↑16.3 |
| Heptatis | 87.9 | 87.2 | 67.5 | 77.3 | 74.2 | 80.6 | 58.1 | 83.8 | 70.9 | 73.5 | 87.5 | 68.7 | 93.3 | 80.0 | 80.0 | 82.7 | ↑5.7 | ↓1.9 | ↑6.00 |
| Housevote | 94.2 | 92.2 | 94.9 | 93.5 | 91.9 | 94.2 | 94.2 | 93.1 | 96.5 | 94.1 | 88.6 | 95.4 | 95.3 | 93.1 | 97.6 | 94.2 | ↑6.1 | ↑5.9 | ↑6.10 |
| Ionosphere | 85.9 | 82.3 | 85.7 | 84.1 | 85.9 | 88.5 | 85.7 | 85.7 | 85.7 | 86.3 | 82.8 | 88.5 | 80.0 | 88.5 | 85.7 | 86.6 | ↑18.0 | ↑8.6 | ↑7.70 |
| Mammographic | 80.3 | 80 | 76.4 | 78.2 | 79.7 | 73.9 | 78.1 | 76.5 | 75.5 | 76.7 | 79.1 | 75.0 | 76.1 | 80.2 | 77.1 | 77.2 | ↑2.40 | ↑0.50 | ↑0.60 |
| Monk-2 | 99.1 | 92.8 | 100 | 96.4 | 100 | 90.1 | 92.7 | 100 | 81.9 | 92.9 | 83.9 | 87.5 | 100 | 89.1 | 85.4 | 90.5 | ↑28.3 | ↑21.8 | ↑18.0 |
| Mushroom | 100 | 99.9 | 99.7 | 99.8 | 97.8 | 99.1 | 100 | 97.9 | 98.7 | 98.7 | 100 | 100 | 99.1 | 100 | 98.7 | 99.4 | ↑1.50 | ↓1.30 | ↓0.60 |
| Phoneme | 81.3 | 78.4 | 80.4 | 79.4 | 81.1 | 80.2 | 81.1 | 77.8 | 78.5 | 79.9 | 83.7 | 77.6 | 78.2 | 79.8 | 79.6 | 79.1 | ↓0.30 | ↓1.00 | ↓1.50 |
| Pima | 75.3 | 77.3 | 75.2 | 76.3 | 77.9 | 77.9 | 72.7 | 77.7 | 76.4 | 76.5 | 79.2 | 84.4 | 66.6 | 76.6 | 76.3 | 76.6 | ↑5.90 | ↑1.90 | ↑1.70 |
| Saheart | 77.4 | 73.2 | 70.9 | 72.1 | 75.2 | 67.7 | 79.3 | 69.5 | 64.1 | 71.2 | 72.3 | 73.9 | 65.2 | 69.5 | 71.7 | 71.4 | ↑1.40 | ↓0.40 | ↑0.90 |
| Thyroid | 85.7 | 83.6 | 78.8 | 81.3 | 80.9 | 78.5 | 85.7 | 85.3 | 80.5 | 82.2 | 66.6 | 71.4 | 95.2 | 66.6 | 60.0 | 76.9 | ↑3.50 | ↑0.00 | ↓4.70 |
| Spambase | 93.8 | 92.5 | 93.5 | 93.0 | 93.3 | 93.5 | 93.1 | 92.8 | 93.8 | 93.3 | 93.4 | 93.1 | 95.2 | 93.4 | 94.1 | 93.7 | ↑0.20 | ↑0.20 | ↑0.50 |
| SPECTHeart | 85.1 | 79.1 | 73.6 | 76.3 | 83.3 | 83.3 | 83.1 | 79.2 | 71.6 | 80.1 | 85.1 | 74.1 | 74.1 | 84.6 | 65.3 | 75.6 | ↑9.30 | ↑11.9 | ↑7.50 |
| Tic-Tac-Toe | 77.1 | 71.8 | 67.2 | 69.5 | 69.7 | 77.6 | 73.9 | 70.6 | 72.2 | 72.8 | 79.2 | 79.2 | 80.2 | 66.6 | 78.9 | 74.1 | ↑8.00 | ↑8.40 | ↑9.40 |
| Titanic | 75.6 | 77.4 | 78.1 | 77.8 | 73.4 | 80.4 | 79.7 | 78.4 | 78.1 | 78.1 | 78.2 | 80.0 | 76.8 | 77.3 | 76.8 | 77.9 | ↑0.30 | ↑1.00 | ↑1.10 |
| Wdbc | 99.1 | 97.8 | 97.1 | 97.5 | 98.2 | 99.1 | 98.2 | 98.2 | 96.4 | 98.1 | 98.2 | 98.2 | 94.7 | 98.2 | 96.4 | 97.7 | ↑1.90 | ↑1.70 | ↑1.30 |
| Wisconsin | 97.2 | 96.5 | 96.5 | 96.5 | 97.1 | 97.8 | 97.1 | 97.1 | 94.2 | 96.7 | 97.1 | 98.5 | 95.7 | 95.7 | 95.6 | 96.1 | ↑1.10 | ↑1.00 | ↑0.40 |

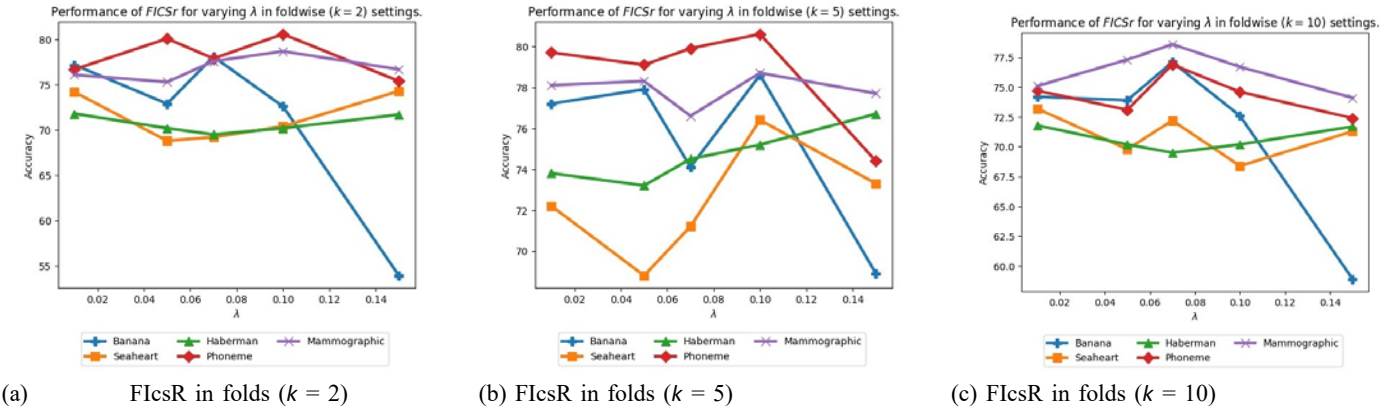

(a) FIcsR in folds (k = 2)  (b) FIcsR in folds (k = 5)  (c) FIcsR in folds (k = 10)

Fig. 2: st-CV and *FIcsR*, Δ accuracy for varying number of batches.

maximum increase in performance of more than 1.11% and upto 5.10% against the nearest competitor One-step, on varying shift levels across all high dimensional datasets. *FIcsR* ability to absorb and retain previous knowledge could be attributed to its better performance. The lag in performance of EIWERM could be due to over flattening the importance weights, while ERM non suitability under different train/test distributions is evident.

**Binary datasets** *FIcsR*, again surpasses state-of-the-art namely RuLISF and One-step. As can be observed from Table II, column $\Delta_2$, *FIcsR* reports a maximum increase in performance of more than 1.74% and up to 5.11% against the nearest competitor One-step, on 4 out of 5 datasets. Though computationally efficient, yet uLSIF and RuLSIF lag in mean accuracy could be due to estimator's sensitivity towards edge examples.

*F. Ablation Study*

We may look at ablating the model by adding more covariate shift. This may be done by increasing the number of fragments to make each fragment represent a smaller distribution support. It may also happen by principled noise injection to perturb the shape of densities.

**Robustness of FIcsR to noise injection** We present results of an ablation study on CIFAR10 and CIFAR100 by adding gaussian noise to the covariates by the pipeline from Lipton et al. [31], with standard deviation (1, 10, 25, 50,75, 100). We observe that *FIcsR* is reasonably robust to different levels of shift. The results are reported in Figure 3b. *FIcsR* shows 89% difference (Δ) at maximum in accuracy for CIFAR10 in comparison to st-cv for noise level (100).

**Effect of more numerous batching** We increase the number of batches while fragmenting the training set. To analyze the robustness of *FIcsR* to the fragmentation of the dataset, we compare *FIcsR* with standard st-CV (**BL2**). The batching results are presented in 3a. We can notice figure 3a results,

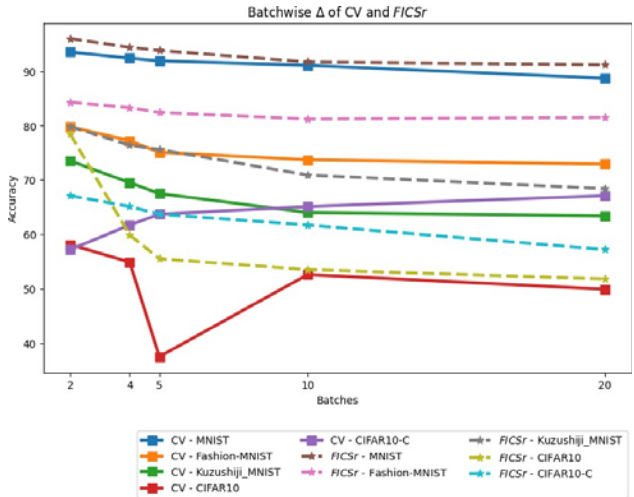

(a) st-CV and *FIcsR*, Δ accuracy for varying number of λ and batches/folds.

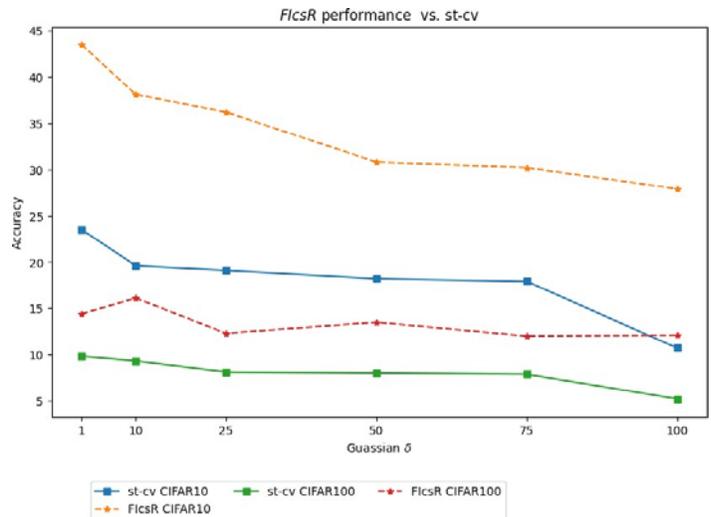

(b) st-CV and *FIcsR*, Δ accuracy for varying number of guassian noise.

where *FIcsR*, Δ is minimum for almost every fragmented dataset batch sequence as compared to st-CV Δ in figure 3a except for CIFAR10-C. *FIcsR* is more robust as compared to st-CV when the number of batches/folds increases. In Table VIII it can be observed that *FIcsR* improves in accuracy with 7.2%, 7.2%, and 2.1% for F-MNIST, K-MNIST, and P-MNIST when batch size is 15% of the training set. For batch size 20% of the training set, *FIcsR* improves 9.7%, 8.1%, and 7.3% for CIFAR100, K-MNIST, and F-MNIST. When the training set is fragmented into 4 batches, *FIcsR* shows 11.3%, 6.9%, 6.1%, and 5.8% improvement in accuracy for CIFAR-100, K-MNIST, F-MNIST, and CIFAR100-C.

**Caliberating penalty** One of our contributions to the paper is introducing the penalty term as above mentioned. We calibrate the penalty term (λ) with different values within the range (0.01, 0.04, 0.07, 0.1) in all of our experiments in batch/fold setup and present the result in Fig 2. We notice that *FIcsR* shows better performance for λ = 0.1 as shown in fig 2b. So, we reported all our experimental results for λ = 0.1. However we expanded the grid with a varying number of folds as well which results are given in fig 2.

## VI. CONCLUSIONS

We propose *FIcsR*, a method for remediating fragmented-induced covariate shift caused by fragmentation of the dataset into a number of sub-datasets (batches or folds) for cross-validation. *FIcsR* sits at the core of cross-validation as the standard tool for tuning hyperparameters in larger data-distributed systems where the entire training set may not be available in the same place at the same time. To summarize *FIcsR*,

1) Demonstrate double mitigation, both with standard and induced-shift.
2) Improves significantly upon state-of-the-art with an increase in average accuracy by more than 5% on both, high dimensional and tabular datasets.
3) Remediates shift under batchwise settings and reports an increase in average accuracy of more than 10% and 25% on no-covariate shift and standard-covariate-shift dataset baselines, respectively.
4) Mitigates shift under foldwise settings and we observe an increase in average accuracy of more than 43% and 28% on image-based and tabular dataset baselines, respectively.
5) Distinguishes itself in complexity as it scales linearly with the size of the training set fragment (batch/fold) only.

TABLE VII: st-CV batch-wise accuracy (complete version of Table III)

| Dataset | Baseline st-CV | Batchwise accuracy | | | | | Mean $\mu_1$ | var $\sigma_1^2$ |
|---|---|---|---|---|---|---|---|---|
| | | $B_1$ | $B_2$ | $B_{\frac{n}{2}}$ | $B_{n-1}$ | $B_n$ | | |
| Training data = 5% , Number_of_Batches = 20 | | | | | | | | |
| MNIST | 94.8 | 89.3 | 87.9 | 89.9 | 88.9 | 88.8 | 88.7 | 0.49 |
| EMNIST | 83.1 | 73.7 | 74.2 | 72.5 | 74 | 70.6 | 72.9 | 1.94 |
| CIFAR-10 | 71.5 | 49.0 | 50.3 | 50.7 | 51.2 | 54.5 | 49.9 | 9.67 |
| CIFAR-100 | 38.2 | 16.5 | 18.1 | 19.3 | 20.4 | 22.7 | 18.3 | 9.17 |
| P-MNIST | 95.1 | 86.1 | 88.9 | 88.7 | 87.2 | 88.3 | 88.4 | 1.41 |
| QMNIST | 75.4 | 63.4 | 62.2 | 66.7 | 58.3 | 63.4 | 63.4 | 5.59 |
| CIFAR10-C | 63.9 | 20.1 | 16.3 | 16.1 | 14.9 | 10.2 | 16.2 | 10.4 |
| CIFAR100-C | 28.8 | 16.3 | 19.1 | 19.7 | 21.6 | 22.3 | 18.4 | 14.2 |
| Training data = 10% , Number_of_Batches = 10 | | | | | | | | |
| MNIST | 94.8 | 91.7 | 91.1 | 90.2 | 89.3 | 91.5 | 91.1 | 1.06 |
| EMNIST | 83.1 | 69.9 | 75.2 | 73.4 | 74.4 | 71.7 | 73.7 | 9.22 |
| CIFAR-10 | 71.5 | 52.8 | 52.9 | 53.7 | 54.5 | 54.1 | 52.6 | 4.87 |
| CIFAR-100 | 38.2 | 20.3 | 22.2 | 23.3 | 24.7 | 21.2 | 21.5 | 5.51 |
| P-MNIST | 95.1 | 92.8 | 91.8 | 91.2 | 91.4 | 91.3 | 91.5 | 0.62 |
| QMNIST | 75.4 | 63.6 | 64.1 | 64.9 | 65.2 | 64.5 | 64.0 | 1.15 |
| CIFAR10- C | 63.9 | 17.6 | 18.4 | 12.2 | 17.4 | 12.9 | 22.6 | 16.8 |
| CIFAR100-C | 28.8 | 21.9 | 25.2 | 26.6 | 27.1 | 20.5 | 22.8 | 16.3 |
| Training data = 20% , Number_of_Batches = 5 | | | | | | | | |
| MNIST | 94.8 | 92.2 | 93.4 | 91.2 | 90.2 | ˇ | 91.9 | 1.59 |
| EMNIST | 83.1 | 74.1 | 72 | 75.3 | 77.2 | ˇ | 75.1 | 4.40 |
| CIFAR-10 | 71.5 | 38.5 | 37.5 | 36.8 | 37.9 | ˇ | 37.5 | 0.50 |
| CIFAR-100 | 38.2 | 22.3 | 23.5 | 24.4 | 24.9 | ˇ | 22.9 | 3.43 |
| P-MNIST | 95.1 | 91.7 | 91.4 | 91.2 | 93.5 | ˇ | 91.8 | 1.01 |
| QMNIST | 75.4 | 69.2 | 67.7 | 65.5 | 66.8 | ˇ | 67.5 | 2.02 |
| CIFAR10- C | 63.9 | 62.0 | 61.8 | 60.7 | 67.5 | ˇ | 62.0 | 9.43 |
| CIFAR100- C | 28.8 | 23.2 | 23.2 | 26.7 | 26.1 | ˇ | 23.8 | 5.77 |
| Training data = 25% , Number_of_Batches = 4 | | | | | | | | |
| MNIST | 94.8 | 93.6 | 92.5 | 91.3 | ˇ | ˇ | 92.4 | 0.92 |
| EMNIST | 83.1 | 77.5 | 78.1 | 78.5 | ˇ | ˇ | 77.2 | 3.12 |
| CIFAR-10 | 71.5 | 52.7 | 56.9 | 58.1 | ˇ | ˇ | 54.9 | 7.02 |
| CIFAR-100 | 38.2 | 22.4 | 23.9 | 24.8 | ˇ | ˇ | 22.8 | 2.7 |
| P-MNIST | 95.1 | 92.2 | 90.5 | 91.6 | ˇ | ˇ | 91.6 | 0.61 |
| QMNIST | 75.4 | 69.2 | 69.5 | 68.8 | ˇ | ˇ | 69.5 | 0.60 |
| CIFAR10- C | 63.9 | 62.1 | 63.1 | 62.6 | ˇ | ˇ | 61.4 | 4.81 |
| CIFAR100- C | 28.8 | 23.4 | 23.8 | 24.5 | ˇ | ˇ | 22.8 | 3.64 |
| Training data = 50% , Number_of_Batches = 2 | | | | | | | | |
| MNIST | 94.8 | 93.7 | ˇ | ˇ | ˇ | ˇ | 93.5 | 0.08 |
| EMNIST | 83.1 | 79.7 | ˇ | ˇ | ˇ | ˇ | 79.8 | 0.04 |
| CIFAR-10 | 71.5 | 60.1 | ˇ | ˇ | ˇ | ˇ | 58.1 | 3.81 |
| CIFAR-100 | 38.2 | 25.4 | ˇ | ˇ | ˇ | ˇ | 24.3 | 1.21 |
| P-MNIST | 95.1 | 93.7 | ˇ | ˇ | ˇ | ˇ | 93.5 | 0.08 |
| QMNIST | 75.4 | 73.3 | ˇ | ˇ | ˇ | ˇ | 73.5 | 0.04 |
| CIFAR10- C | 63.9 | 65.5 | ˇ | ˇ | ˇ | ˇ | 64.3 | 1.32 |
| CIFAR100- C | 28.8 | 27.5 | ˇ | ˇ | ˇ | ˇ | 26.4 | 1.21 |

TABLE VIII: *FIcsR* Batchwise (complete version of Table IV)

| Dataset | Batchwise accuracy | | | | | | Mean | var | $\Delta_3 = \mu_2 - \mu_1$ |
|---|---|---|---|---|---|---|---|---|---|
| | $B_1$ | $B_2$ | $B_3$ | $B_{\frac{n}{2}}$ | $B_{n-1}$ | $B_n$ | $\mu_2$ | $\sigma_2^2$ | $\Delta_3(\%)$ |
| Training data = 5% , Number_of_Batches = 20 | | | | | | | | | |
| MNIST | 90.7 | 90.6 | 91 | 91.7 | 91.4 | 91.8 | 91.2 | 0.09 | ↑ 2.81 |
| EMNIST | 81.5 | 81.7 | 81.2 | 81.4 | 81.5 | 81.9 | 81.5 | 0.058 | ↑ 11.7 |
| CIFAR-10 | 50.9 | 51.4 | 52.2 | 48.9 | 50.3 | 57.4 | 51.8 | 7.18 | ↑ 3.81 |
| CIFAR-100 | 23.9 | 18.2 | 18.5 | 17.8 | 23.9 | 18.3 | 20.1 | 7.26 | ↑ 9.83 |
| P-MNIST | 91.1 | 89.8 | 90.2 | 90.4 | 91.5 | 90.7 | 90.3 | 0.26 | ↑ 2.14 |
| QMNIST | 68.5 | 69.4 | 67.2 | 68 | 67.8 | 66.9 | 68.4 | 0.63 | ↑ 7.88 |
| CIFAR10-C | 46.4 | 54.3 | 57.8 | 61.1 | 61.5 | 61.8 | 57.2 | 30.1 | ↑ 253 |
| CIFAR100-C | 11.9 | 17.2 | 18.5 | 21.3 | 22.1 | 24.9 | 19.3 | 17.1 | ↑ 4.89 |
| Training data = 10% , Number_of_Batches = 10 | | | | | | | | | |
| MNIST | 91.9 | 91.7 | 91.2 | 91.8 | 91.3 | 91.8 | 91.7 | 0.08 | ↑ 0.65 |
| EMNIST | 79.5 | 82.4 | 81.6 | 79.5 | 82.3 | 81.9 | 81.2 | 1.21 | ↑ 10.1 |
| CIFAR-10 | 52.1 | 53.1 | 48.5 | 59.3 | 52.5 | 55.7 | 53.5 | 11.09 | ↑ 1.71 |
| CIFAR-100 | 27.2 | 25.8 | 20.4 | 17.0 | 21.9 | 22.8 | 22.5 | 11.3 | ↑ 4.65 |
| P-MNIST | 91.6 | 91.9 | 91.3 | 91.6 | 90.1 | 91.2 | 91.5 | 0.31 | 0.00 |
| QMNIST | 71.4 | 70.4 | 71.7 | 70.7 | 70.5 | 70.9 | 70.9 | 0.71 | ↑ 10.7 |
| CIFAR10-C | 52.7 | 59.9 | 61.9 | 64.4 | 66.1 | 65.7 | 61.7 | 21.1 | ↑ 173 |
| CIFAR100-C | 16.2 | 22.1 | 24.8 | 27.2 | 26.8 | 27.3 | 21.1 | 15.6 | ↓ 7.45 |
| Training data = 20% , Number_of_Batches = 5 | | | | | | | | | |
| MNIST | 93.6 | 93.8 | 94.3 | 93.7 | 93.8 | ˇ | 93.8 | 0.07 | ↑ 2.06 |
| EMNIST | 82.8 | 83.1 | 82.1 | 81.4 | 82.6 | ˇ | 82.4 | 0.44 | ↑ 6.73 |
| CIFAR-10 | 50.3 | 56.2 | 53.5 | 57.8 | 59.9 | ˇ | 55.5 | 11.3 | ↑ 48.0 |
| CIFAR-100 | 34.2 | 35.4 | 33.9 | 34.9 | 34.7 | ˇ | 34.6 | 11.7 | ↑ 51.1 |
| P-MNIST | 94.1 | 93.9 | 93.6 | 94.1 | 94.3 | ˇ | 94 | 0.07 | ↑ 0.53 |
| QMNIST | 75.4 | 76.3 | 75.8 | 75.1 | 75.6 | ˇ | 75.6 | 0.21 | ↑ 12.0 |
| CIFAR10-C | 58.4 | 63.3 | 64.3 | 66.5 | 66.2 | ˇ | 63.7 | 8.53 | ↓ 3.80 |
| CIFAR100-C | 22.1 | 25.4 | 27.4 | 28.9 | 29.7 | ˇ | 26.7 | 7.43 | ↑ 12.1 |
| Training data = 25% , Number_of_Batches = 4 | | | | | | | | | |
| MNIST | 94.4 | 94.4 | 94.3 | 94.4 | ˇ | ˇ | 94.4 | 0.003 | ↑ 2.16 |
| EMNIST | 82.8 | 83.2 | 83.6 | 83.5 | ˇ | ˇ | 83.3 | 0.13 | ↑ 4.38 |
| CIFAR-10 | 56.8 | 57.3 | 62.2 | 63.4 | ˇ | ˇ | 59.9 | 8.47 | ↑ 9.10 |
| CIFAR-100 | 33.9 | 34.1 | 34.7 | 33.5 | ˇ | ˇ | 34.1 | 0.18 | ↑ 49.5 |
| P-MNIST | 94.4 | 94.3 | 94.5 | 94.5 | ˇ | ˇ | 94.4 | 0.009 | ↑ 3.05 |
| QMNIST | 77.2 | 75.5 | 77.6 | 75.3 | ˇ | ˇ | 76.4 | 1.37 | ↑ 11.0 |
| CIFAR10-C | 60.9 | 64.4 | 66.8 | 68.3 | ˇ | ˇ | 65.1 | 7.81 | ↑ 6.02 |
| CIFAR100-C | 24.7 | 28.2 | 29.7 | 31.9 | ˇ | ˇ | 28.6 | 6.86 | ↑ 25.4 |
| Training data = 50% , Number_of_Batches = 2 | | | | | | | | | |
| MNIST | 95.9 | 96.1 | ˇ | ˇ | ˇ | ˇ | 96 | 0.02 | ↑ 2.67 |
| EMNIST | 84.2 | 84.4 | ˇ | ˇ | ˇ | ˇ | 84.3 | 0.02 | ↑ 5.63 |
| CIFAR-10 | 76.3 | 80.6 | ˇ | ˇ | ˇ | ˇ | 78.4 | 4.62 | ↑ 34.9 |
| CIFAR-100 | 39.8 | 39.9 | ˇ | ˇ | ˇ | ˇ | 39.85 | .002 | ↑ 63.9 |
| P-MNIST | 95.7 | 96.1 | ˇ | ˇ | ˇ | ˇ | 95.9 | 0.08 | ↑ 2.56 |
| QMNIST | 79.3 | 80.4 | ˇ | ˇ | ˇ | ˇ | 79.8 | 0.61 | ↑ 8.57 |
| CIFAR10-C | 65.5 | 68.6 | ˇ | ˇ | ˇ | ˇ | 67.1 | 2.4 | ↑ 4.35 |
| CIFAR100-C | 31.2 | 34.8 | ˇ | ˇ | ˇ | ˇ | 33 | 3.24 | ↑ 25.0 |

TABLE IX: st-CV foldchwise Accuracy

| Dataset | Baseline | | Foldwise accuracy | | | | | Mean | var |
|---|---|---|---|---|---|---|---|---|---|
| | st-CV | FIcsR | $k_1$ | $k_2$ | $k_{\frac{n}{2}}$ | $k_{n-1}$ | $k_n$ | $\mu_9$ | $\sigma_3^2$ |
| **Number_of_Folds = 10** | | | | | | | | | |
| MNIST | 94.8 | 97.9 | 95.1 | 94.8 | 94.1 | 95.3 | 94.6 | 94.9 | 0.17 |
| P-MNIST | 95.1 | 97.6 | 94.1 | 94.5 | 94.3 | 94.3 | 93.5 | 94.0 | 0.11 |
| EMNIST | 83.1 | 88.4 | 83.9 | 84.2 | 82.6 | 84.2 | 83.7 | 83.7 | 0.35 |
| QMNIST | 75.4 | 89.2 | 85.5 | 85.6 | 86.2 | 86.6 | 86.2 | 85.9 | 0.17 |
| KanadaMNIST | 97.7 | 99.2 | 94.1 | 95.1 | 96.2 | 95.8 | 96.9 | 95.9 | 0.92 |
| CIFAR-10 | 71.5 | 88.7 | 65.6 | 66.4 | 64.4 | 65.3 | 63.6 | 64.4 | 0.94 |
| CIFAR-100 | 38.2 | 58.7 | 31.4 | 29.9 | 30.2 | 32.2 | 30.7 | 31.8 | 0.69 |
| CIFAR10-C | 63.9 | 73.3 | 66.6 | 67.5 | 65.5 | 66.1 | 67.7 | 66.4 | 0.69 |
| CIFAR100-C | 28.8 | 39.4 | 33.4 | 32.8 | 32.8 | 32.1 | 31.9 | 32.6 | 0.29 |
| SVHN | 85.1 | 90.8 | 87.3 | 84.6 | 19.2 | 84.4 | 86.2 | 79.1 | 707 |
| Caltech101 | 52.3 | 51.7 | 54.8 | 58.6 | 53.5 | 52.9 | 55.7 | 55.7 | 4.02 |
| Tiny-ImageNet | 42.5 | 59.2 | 41.1 | 39.3 | 33.9 | 40.8 | 38.2 | 38.6 | 6.76 |
| STL-10 | 43.6 | 56.2 | 45.2 | 42.7 | 40.1 | 39.9 | 42.5 | 42.5 | 3.79 |
| **Number_of_Folds = 5** | | | | | | | | | |
| MNIST | 94.8 | 97.9 | 94.7 | 94.3 | 95.1 | 95.1 | 94.9 | 94.8 | 0.09 |
| P-MNIST | 95.1 | 97.6 | 93.5 | 94.1 | 93.3 | 94.0 | | 93.6 | 0.09 |
| EMNIST | 83.1 | 88.4 | 84.1 | 83.8 | 84.3 | 83.5 | 84.7 | 84.1 | 0.17 |
| QMNIST | 75.4 | 89.2 | 86.8 | 84.3 | 86.6 | 85.8 | 85.7 | 85.8 | 0.77 |
| KanadaMNIST | 97.7 | 99.2 | 93.1 | 94.4 | 94.5 | 96.2 | 95.9 | 94.8 | 1.26 |
| CIFAR-10 | 71.5 | 88.7 | 59.7 | 63.4 | 61.3 | 61.6 | 62.7 | 61.8 | 1.61 |
| CIFAR-100 | 38.2 | 58.7 | 29.8 | 31.3 | 31.9 | 33.5 | 31.1 | 31.5 | 1.44 |
| CIFAR10-C | 63.9 | 73.3 | 64.9 | 65.8 | 65.1 | 66.2 | 65.1 | 65.4 | 0.25 |
| CIFAR100-C | 28.8 | 39.4 | 32.3 | 32.3 | 32.0 | 31.5 | 31.5 | 31.9 | 0.13 |
| SVHN | 85.1 | 90.8 | 83.0 | 84.2 | 84.9 | 86.2 | 85.7 | 84.8 | 1.28 |
| Caltech101 | 52.3 | 51.7 | 55.9 | 52.3 | 56.4 | 50.7 | 56.2 | 54.3 | 5.51 |
| Tiny-ImageNet | 42.5 | 59.2 | 41.8 | 33.7 | 39.3 | 40.5 | 34.8 | 51.1 | 10.2 |
| STL-10 | 43.6 | 56.2 | 44.8 | 41.5 | 46.5 | 41.7 | 43.7 | 43.6 | 3.57 |
| **Number_of_Folds = 2** | | | | | | | | | |
| MNIST | 94.8 | 97.9 | 91.4 | 92.4 | | | | 93.7 | 0.25 |
| P-MNIST | 95.1 | 97.6 | 92.5 | 92.4 | | | | 92.4 | .003 |
| EMNIST | 83.1 | 88.4 | 83.9 | 84.2 | | | | 84.1 | 0.02 |
| QMNIST | 75.4 | 89.2 | 82.5 | 84.4 | | | | 83.4 | 0.91 |
| KanadaMNIST | 97.7 | 99.2 | 93.1 | 94.1 | | | | 93.6 | 0.25 |
| CIFAR-10 | 71.5 | 88.7 | 57.3 | 58.8 | | | | 58.1 | 0.56 |
| CIFAR-100 | 38.2 | 58.7 | 26.3 | 25.7 | | | | 26.1 | 0.09 |
| CIFAR10-C | 63.9 | 73.3 | 63.1 | 64.9 | | | | 64.2 | 0.81 |
| CIFAR100-C | 28.8 | 39.4 | 28.2 | 28.1 | | | | 28.1 | .003 |
| SVHN | 85.1 | 90.8 | 82.1 | 19.0 | | | | 50.5 | 995 |
| Caltech101 | 52.3 | 51.7 | 50.9 | 49.3 | | | | 50.1 | 0.64 |
| Tiny-ImageNet | 42.5 | 59.2 | 39.6 | 32.8 | | | | 36.2 | 11.5 |
| STL-10 | 43.6 | 56.2 | 33.8 | 39.9 | | | | 36.8 | 18.6 |

TABLE X: *FIcsR* foldwise Accuracy

| Dataset | Baseline | | Foldwise accuracy | | | | | Mean | var | $\Delta_4 = \mu_9 - \mu_{10}$ |
|---|---|---|---|---|---|---|---|---|---|---|
| | st-CV | FIcsR | $k_1$ | $k_2$ | $k_{\frac{n}{2}}$ | $k_{n-1}$ | $k_n$ | $\mu_{10}$ | $\sigma_4^2$ | $\Delta_4$(%) |
| **Number_of_Folds = 10** | | | | | | | | | | |
| MNIST | 94.8 | 97.9 | 97.8 | 97.3 | 97.4 | 97.2 | 97.4 | 97.4 | 0.04 | ↑ 2.50 |
| P-MNIST | 95.1 | 97.6 | 91.2 | 91.1 | 91.3 | 91.2 | 90.9 | 91.1 | 0.02 | ↓ 2.90 |
| EMNIST | 83.1 | 88.4 | 88.9 | 88.5 | 89.1 | 89.5 | 88.5 | 88.9 | 0.14 | ↑ 5.20 |
| QMNIST | 75.4 | 89.2 | 94.8 | 94.7 | 94.8 | 95.2 | 94.8 | 94.8 | 0.03 | ↑ 8.90 |
| KanadaMNIST | 97.7 | 99.2 | 97.8 | 97.7 | 97.5 | 97.7 | 97.6 | 97.6 | 0.01 | ↑ 1.70 |
| CIFAR-10 | 71.5 | 88.7 | 85.7 | 85.4 | 85.5 | 85.4 | 85.3 | 85.4 | 0.02 | ↑ 21.0 |
| CIFAR-100 | 38.2 | 58.7 | 63.3 | 63.8 | 64.3 | 62.9 | 39.2 | 58.7 | 95.2 | ↑ 26.9 |
| CIFAR10-C | 63.9 | 73.3 | 71.1 | 69.9 | 71.3 | 70.5 | 69.2 | 70.4 | 0.6 | ↑ 4.00 |
| CIFAR100-C | 28.8 | 39.4 | 32.7 | 22.8 | 38.5 | 33.0 | 35.1 | 32.4 | 27.5 | ↓ 0.20 |
| SVHN | 85.1 | 90.8 | 95.6 | 95.9 | 95.8 | 95.4 | 95.5 | 95.6 | .034 | ↑ 16.5 |
| Caltech101 | 52.3 | 51.7 | 95.6 | 95.3 | 95.9 | 95.7 | 60.4 | 88.5 | 198 | ↑ 32.8 |
| Tiny-ImageNet | 42.5 | 59.2 | 57.9 | 56.7 | 49.1 | 53.4 | 57.5 | 54.8 | 10.9 | ↑ 16.2 |
| STL-10 | 43.6 | 56.2 | 51.3 | 50.6 | 52.7 | 49.4 | 47.2 | 50.2 | 4.33 | ↑7.70 |
| **Number_of_Folds = 5** | | | | | | | | | | |
| MNIST | 94.8 | 97.9 | 97.5 | 97.3 | 97.1 | 97.3 | 97.2 | 97.2 | .017 | ↑ 2.40 |
| P-MNIST | 95.1 | 97.6 | 97.5 | 97.5 | 97.6 | 97.6 | 97.5 | 97.54 | .002 | ↑ 3.94 |
| EMNIST | 83.1 | 88.4 | 88.3 | 88.5 | 88.6 | 88.7 | 88.8 | 88.5 | .029 | ↑ 4.40 |
| QMNIST | 75.4 | 89.2 | 94.6 | 94.7 | 94.7 | 94.7 | 95.1 | 94.7 | 0.03 | ↑ 8.90 |
| KanadaMNIST | 97.7 | 99.2 | 97.2 | 97.3 | 97.3 | 97.4 | 97.4 | 97.3 | .005 | ↑2.50 |
| CIFAR-10 | 71.5 | 88.7 | 71.43 | 71.65 | 72.18 | 72.27 | 73.47 | 72.2 | .502 | ↑ 10.4 |
| CIFAR-100 | 38.2 | 58.7 | 67.8 | 68.1 | 67.7 | 67.9 | 39.2 | 62.1 | 132 | ↑30.6 |
| CIFAR10-C | 63.9 | 73.3 | 91.2 | 91.5 | 91.2 | 91.4 | 69.6 | 86.9 | 75.5 | ↑21.5 |
| CIFAR100-C | 28.8 | 39.4 | 27.5 | 38.1 | 28.8 | 31.4 | 33.0 | 31.7 | 13.7 | ↓ 0.20 |
| SVHN | 85.1 | 90.8 | 96.9 | 97.1 | 96.8 | 96.7 | 92.9 | 96.1 | 2.54 | ↑11.3 |
| Caltech101 | 52.3 | 51.7 | 94.9 | 95.1 | 94.2 | 94.6 | 57.6 | 87.2 | 220 | ↑32.9 |
| Tiny-ImageNet | 42.5 | 59.2 | 56.3 | 51.7 | 53.2 | 52.8 | 38.3 | 50.5 | 39.3 | ↓0.60 |
| STL-10 | 43.6 | 56.2 | 52.9 | 48.7 | 50.6 | 53.3 | 50.9 | 51.2 | 3.49 | ↑ 7.50 |
| **Number_of_Folds = 2** | | | | | | | | | | |
| MNIST | 94.8 | 97.9 | 96.6 | 96.7 | ˘ | ˘ | ˘ | 96.6 | .002 | ↑ 2.90 |
| P-MNIST | 95.1 | 97.6 | 96.9 | 96.9 | ˘ | ˘ | ˘ | 96.9 | 0.00 | ↑4.50 |
| EMNIST | 83.1 | 88.4 | 87.1 | 87.0 | ˘ | ˘ | ˘ | 87.05 | .002 | ↑2.95 |
| QMNIST | 75.4 | 89.2 | 93.7 | 93.6 | ˘ | ˘ | ˘ | 93.6 | .002 | ↑10.2 |
| KanadaMNIST | 97.7 | 99.2 | 96.7 | 96.7 | ˘ | ˘ | ˘ | 96.7 | 0.00 | ↑3.10 |
| CIFAR-10 | 71.5 | 88.7 | 69.2 | 67.9 | ˘ | ˘ | ˘ | 68.5 | 0.43 | ↑10.4 |
| CIFAR-100 | 38.2 | 58.7 | 55.2 | 32.6 | ˘ | ˘ | ˘ | 43.9 | 128 | ↑17.8 |
| CIFAR10-C | 63.9 | 73.3 | 91.3 | 66.5 | ˘ | ˘ | ˘ | 78.9 | 154 | ↑14.7 |
| CIFAR100-C | 28.8 | 39.4 | 39.3 | 22.8 | ˘ | ˘ | ˘ | 31.1 | 68.1 | ↑3.00 |
| SVHN | 85.1 | 90.8 | 96.0 | 91.5 | | | | 93.7 | 5.06 | ↑43.2 |
| Caltech101 | 52.3 | 51.7 | 94.7 | 53.4 | ˘ | ˘ | ˘ | 74.1 | 426 | ↑24.0 |
| Tiny-ImageNet | 42.5 | 59.2 | 55.2 | 45.7 | ˘ | ˘ | ˘ | 50.4 | 22.5 | ↑ 14.2 |
| STL-10 | 43.6 | 56.2 | 43.1 | 44.8 | ˘ | ˘ | ˘ | 43.9 | 1.44 | ↑ 6.10 |

## REFERENCES

[1] J. Alcalá-Fdez, A. Fernández, J. Luengo, J. Derrac, S. García, L. Sánchez, and F. Herrera, "Keel data-mining software tool: data set repository, integration of algorithms and experimental analysis framework." *Journal of Multiple-Valued Logic & Soft Computing*, vol. 17, 2011.

[2] S. Bickel, M. Brückner, and T. Scheffer, "Discriminative learning for differing training and test distributions," in *Proceedings of the 24th international conference on Machine learning*, 2007, pp. 81–88.

[3] W. Bukaew and S. Yoo-Kong, "One-parameter generalised fisher information," *arXiv:2107.10578*, 2021.

[4] T. Clanuwat, M. Bober-Irizar, A. Kitamoto, A. Lamb, K. Yamamoto, and D. Ha, "Deep learning for classical japanese literature," *arXiv:1812.01718*, 2018.

[5] A. Coates, A. Ng, and H. Lee, "An analysis of single-layer networks in unsupervised feature learning," in *Proceedings of the fourteenth international conference on artificial intelligence and statistics*. JMLR Workshop and Conference Proceedings, 2011, pp. 215–223.

[6] G. Cohen, S. Afshar, J. Tapson, and A. V. Schaik, "Emnist: Extending mnist to handwritten letters," *2017 International Joint Conference on Neural Networks (IJCNN)*, 2017.

[7] C. Cortes, M. Mohri, M. Riley, and A. Rostamizadeh, "Sample selection bias correction theory," in *Algorithmic Learning Theory: 19th International Conference, ALT 2008, Budapest, Hungary, October 13-16, 2008. Proceedings 19*. Springer, 2008, pp. 38–53.

[8] H. Cramer, "Mathematical methods of statistics, princeton, 1946," *Math Rev (Math-SciNet) MR16588 Zentralblatt MATH*, vol. 63, p. 300, 1946.

[9] Z. Fan, "Statistics 200: Introduction to statistical inference," Lecture, 2016, stanford University, Autumn.

[10] T. Fang, N. Lu, G. Niu, and M. Sugiyama, "Rethinking importance weighting for deep learning under distribution shift," *Advances in neural information processing systems*, vol. 33, pp. 11 996–12 007, 2020.

[11] L. Fei-Fei, R. Fergus, and P. Perona, "Learning generative visual models from few training examples: An incremental bayesian approach tested on 101 object categories," *Computer Vision and Pattern Recognition Workshop*, 2004.

[12] D. Hendrycks and T. Dietterich, "Benchmarking neural network robustness to common corruptions and perturbations," *Proceedings of the International Conference on Learning Representations*, 2019.

[13] J. Huang, A. Gretton, K. Borgwardt, B. Schölkopf, and A. Smola, "Correcting sample selection bias by unlabeled data," *Advances in neural information processing systems*, vol. 19, 2006.

[14] S. Jirayucharoensak, P.-N. Setha, and P. Israsena, "Eeg-based emotion recognition using deep learning network with principal component based covariate shift adaptation," *The Scientific World Journal*, vol. 1, pp. 1200–1225, 2014.

[15] T. Kanamori and M. Sugiyama, "Application of covariate shift adaptation techniques in brain–computer interfaces," *IEEE Transactions on Biomedical Engineering*, vol. 57, pp. 1318–1324, 2010.

[16] A. Krizhevsky, G. Hinton *et al.*, "Learning multiple layers of features from tiny images," 2009.

[17] M. Kukar, "Transductive reliability estimation for medical diagnosis," *Artificial Intelligence in Medicine*, vol. 29, no. 1-2, pp. 81–106, 2003.

[18] S. Kullback and R. A. Leibler, "On information and sufficiency," *The annals of mathematical statistics*, vol. 22, no. 1, pp. 79–86, 1951.

[19] Y. LeCun, "The mnist database of handwritten digits," *http://yann.lecun.com/exdb/mnist/*, 1998.

[20] E. L. Lehmann and G. Casella, *Theory of point estimation*. Springer Science & Business Media, 2006.

[21] Y. Makoto, M. Sugiyama, and M. Tomoko, "Semi-supervised speaker identification under covariate shift," *Signal Processing*, vol. 90, pp. 2353–2361, 2010.

[22] R. Martin, "Lecture notes on advanced statistical theory," *Supplement to the lectures for Stat*, vol. 511, 2016.

[23] S. Masashi and M. Klaus-Robert, "Input-dependent estimation of generalization error under covariate shift," *Statistics & Risk Modeling*, vol. 23, no. 4/2005, pp. 249–279, 2005.

[24] J. G. Moreno-Torres, J. A. Sáez, and F. Herrera, "Study on the impact of partition-induced dataset shift on $k$-fold cross-validation," *IEEE Transactions on Neural Networks and Learning Systems*, vol. 23, no. 8, pp. 1304–1312, 2012.

[25] N. Mu and J. Gilmer, "Mnist-c: A robustness benchmark for computer vision," *arXiv:1906.02337*, 2019.

[26] Y. Netzer, T. Wang, A. Coates, A. Bissacco, B. Wu, and A. Y. Ng, "Reading digits in natural images with unsupervised feature learning," 2011.

[27] T. Nishiyama, "A new lower bound for kullback-leibler divergence based on hammersley-chapman-robbins bound," *arXiv:1907.00288*, 2019.

[28] R. Pascanu and Y. Bengio, "Revisiting natural gradient for deep networks," *arXiv:1301.3584*, 2013.

[29] V. U. Prabhu, "Kannada-mnist: A new handwritten digits dataset for the kannada language," *arXiv preprint arXiv:1908.01242*, 2019.

[30] J. Quiñonero-Candela, M. Sugiyama, A. Schwaighofer, and N. D. Lawrence, *Dataset shift in machine learning*. Mit Press, 2008.

[31] S. Rabanser, S. Günnemann, and Z. Lipton, "Failing loudly: An empirical study of methods for detecting dataset shift," *Advances in Neural Information Processing Systems*, vol. 32, 2019.

[32] C. Rudin and B. Ustun, "Optimized scoring systems: Toward trust in machine learning for healthcare and criminal justice," *Interfaces*, vol. 48, no. 5, pp. 449–466, 2018.

[33] B. Schölkopf, A. J. Smola, F. Bach *et al.*, *Learning with kernels: support vector machines, regularization, optimization, and beyond*. MIT press, 2002.

[34] L. A. Shepp, "Radon-nikodym derivatives of gaussian measures," *The Annals of Mathematical Statistics*, pp. 321–354, 1966.

[35] H. Shimodaira, "Improving predictive inference under covariate shift by weighting the log-likelihood function," *Journal of statistical planning and inference*, vol. 90, no. 2, pp. 227–244, 2000.

[36] M. Sugiyama and M. Kawanabe, *Machine learning in non-stationary environments: Introduction to covariate shift adaptation*. MIT press, 2012.

[37] M. Sugiyama, M. Krauledat, and K.-R. Müller, "Covariate shift adaptation by importance weighted cross validation." *Journal of Machine Learning Research*, vol. 8, no. 5, 2007.

[38] M. Sugiyama, T. Suzuki, and T. Kanamori, *Density ratio estimation in machine learning*. Cambridge University Press, 2012.

[39] V. N. Vapnik, V. Vapnik *et al.*, "Statistical learning theory," 1998.

[40] F. Wilcoxon, "Individual comparisons by ranking methods," in *Breakthroughs in statistics: Methodology and distribution*. Springer, 1992, pp. 196–202.

[41] A. Wong, A. Hryniowski, and X. Y. Wang, "Insights into fairness through trust: multi-scale trust quantification for financial deep learning," *arXiv:2011.01961*, 2020.

[42] H. Xiao, K. Rasul, and R. Vollgraf, "Fashion-mnist: a novel image dataset for benchmarking machine learning algorithms," *arXiv:1708.07747*, 2017.

[43] C. Yadav and L. Bottou, "Cold case: The lost mnist digits," *Advances in neural information processing systems*, vol. 32, 2019.

[44] M. Yamada, T. Suzuki, T. Kanamori, H. Hachiya, and M. Sugiyama, "Relative density-ratio estimation for robust distribution comparison," *Neural computation*, vol. 25, no. 5, pp. 1324–1370, 2013.

[45] B. Zadrozny, "Learning and evaluating classifiers under sample selection bias," in *Proceedings of the twenty-first international conference on Machine learning*, 2004, p. 114.

[46] T. Zhang, I. Yamane, N. Lu, and M. Sugiyama, "A one-step approach to covariate shift adaptation," in *Asian Conference on Machine Learning*. PMLR, 2020, pp. 65–80.

[47] Y.-J. Zhang, Z.-Y. Zhang, P. Zhao, and M. Sugiyama, "Adapting to continuous covariate shift via online density ratio estimation," *Advances in Neural Information Processing Systems*, vol. 36, 2024.